\pdfoutput=1

\documentclass[11pt]{article}
\usepackage{authblk}
\usepackage[]{acl}

\usepackage{times}
\usepackage{latexsym}

\usepackage[T1]{fontenc}

\usepackage[utf8]{inputenc}

\usepackage{microtype}

\usepackage{graphicx}
\graphicspath{pics/}
\usepackage{caption}
\usepackage{boldline} 
\usepackage{multirow}
\usepackage{sidecap}
\usepackage{subcaption}
\usepackage{natbib}
\usepackage{amsmath}
\usepackage{amsfonts}

\usepackage{xcolor,colortbl}
\usepackage{bm}
\usepackage{hyperref}
\usepackage{rotating}
\usepackage{adjustbox}
\usepackage{algorithm}
\usepackage{algpseudocode} 
\usepackage{array}
\definecolor{green}{rgb}{0.0, 0.5, 0.0}
\definecolor{red}{rgb}{0.82, 0.1, 0.26}
\definecolor{codegray}{gray}{0.9}

\newlength{\Oldarrayrulewidth}

\usepackage{booktabs}

\usepackage{listings}
\definecolor{codegreen}{rgb}{0,0.6,0}
\definecolor{codegray}{rgb}{0.5,0.5,0.5}
\definecolor{codepurple}{rgb}{0.58,0,0.82}
\definecolor{backcolour}{rgb}{0.95,0.95,0.92}
\lstdefinestyle{mystyle}{
  backgroundcolor=\color{backcolour}, commentstyle=\color{codegreen},
  keywordstyle=\color{magenta},
  numberstyle=\tiny\color{codegray},
  stringstyle=\color{codepurple},
  basicstyle=\ttfamily\footnotesize,
  breakatwhitespace=false,         
  breaklines=true,                 
  captionpos=b,                    
  keepspaces=true,                 
  numbers=left,                    
  numbersep=5pt,                  
  showspaces=false,                
  showstringspaces=false,
  showtabs=false,                  
  tabsize=2
}
\title{ Can Pretrained Language Models Derive Correct Semantics from Corrupt Subwords under Noise? }  

\author{\textbf{Xinzhe Li}, \textbf{Ming Liu}, \textbf{Shang Gao}}
\affil{School of IT, Deakin University, Australia \\
\texttt{\{lixinzhe, m.liu,shang.gao\}@deakin.edu.au}}
\begin{document}
\maketitle

\begin{abstract}
For Pretrained Language Models (PLMs), their susceptibility to noise has recently been linked to subword segmentation. However, it is unclear which aspects of segmentation affect their understanding. This study assesses the robustness of PLMs against various disrupted segmentation caused by noise. An evaluation framework for subword segmentation, named Contrastive Lexical Semantic (CoLeS) probe, is proposed. It provides a systematic categorization of segmentation corruption under noise and evaluation protocols by generating contrastive datasets with canonical-noisy word pairs.  
Experimental results indicate that PLMs are unable to accurately compute word meanings if the noise introduces completely different subwords, small subword fragments, or a large number of additional subwords, particularly when they are inserted within other subwords.
\end{abstract}

\section{Introduction}
\begin{table*}

    \centering
    \small
    \begin{tabular}{l|l|r|r|r}
    \toprule
    \multirow{2}{*}{Corruption Types} & \multirow{2}{*}{Examples} & \multicolumn{3}{c}{Segmentation Sets} \\
    \cmidrule{3-5}
    & & Missing & Overlap & Additive \\
    \midrule
    Complete (intact)
    & tasty $\rightarrow$ taaasty 
    & tasty & & ta, aa, sty\\
    
    Complete 
    & stun $\rightarrow$ stunn
    & s, tun & & stu, nn \\
    
    Partial  
    & effectiveness $\rightarrow$ efeectiveness & effect & iveness & efe, ect \\
    
    Additive (infix)
    & insubstantial $\rightarrow$ insuubstantial & & ins, ub, stan, tial & u \\
    
    Additive (affix)
    & hilarious $\rightarrow$ hilariousss & & hil, ario, us & s, s \\
    
    Missing  
    & insubstantial $\rightarrow$ insstantial & ub & ins, stan, tial & \\
    \bottomrule
    \end{tabular}
    \caption{Examples of different types of segmentation corruption. Complete/partial: completely/partially disrupting the original segmentation;  additive: creating unnecessary subwords; missing: ignoring a token.  A distinct form of complete corruption, referred to as ``intact corruption'', arises when a clean word is tokenized into a single subword that does not appear in the segmentation of its noisy counterpart. In the given example of intact corruption, the term ``tasty'' serves as an intact token.}
    \label{tab:subword_sets}
\end{table*}

The capability to understand the meaning of noisy words through character arrangements is a crucial aspect of human cognitive abilities \citep{rawlinson2007letter}. This capability is highly sought after in practical applications such as machine translation and sentiment analysis \citep{belinkov2017synthetic}. However, despite their success in in-distribution test data with standardized word forms, Pretrained Language Models (PLMs), which serve as the backbone models, tend to perform poorly on rare or noisy words \citep{kumar-etal-2020-noisy,baron2015words}. These noisy words may be caused by accidental typos \citep{belinkov2017synthetic} or spelling variants on social media \citep{ritter-etal-2010-unsupervised}.

Prior studies show that most subword-based PLMs perform poorly under noise largely due to subword segmentation \citep{zhuang2022typos}, while character-based PLMs show more robustness \citep{el-boukkouri-etal-2020-characterbert}. Examining the impact of subword segmentation factors on PLMs is also crucial for defending against the adversarial attacks that leverage the sensitivity of subword segmentation to noise \citep{liu-etal-2022-character}. However, rare work has investigated how the subword segmentation from noisy words affects the word meaning. 

To help address this question, we design and develop a contrastive framework (CoLes) to assess the robustness of PLMs in the face of various forms of segmentation corruption. 
As subword segmentation can be influenced by noise in various ways, such as adding extra subwords or losing original subwords, we systematically categorize the ways into four main categories and two additional sub-categories based on three subword sets, as exemplified in Table \ref{tab:subword_sets}. Two types of noise models are proposed to effectively generate all the types of corruption except missing corruption, and a contrastive dataset consisting of noisy and standard word pairs is created. This framework enables us to evaluate the significance of preserved subwords and the impact of subwords added by noise.

The experimental results provide the following insights: 1) complete corruption: the PLMs struggle to infer meaning accurately if no subwords from the original segmentation are retained. The worst performance is observed when the meaning of original words is stored in the embedding; 2) partial corruption: preserving larger subword chunks can aid the understanding of PLMs, whereas retaining smaller subword pieces tend to be ineffective; and 3) additive corruption: even with all original subwords, however, the addition of subwords can harm the meaning of words, particularly when they are placed within other subwords. The more additive subwords, the greater the deviation in word semantics.
All the results are consistent on the three PLMs with different vocabularies and segmentation algorithms.


\section{Contrastive Lexical-Semantic Probe} \label{sec:coles}
The CoLeS probe framework has segmentation corruption and noise models that produce noisy words leading to different types of segmentation corruption. These noisy words, along with their corresponding canonical forms, are organized in a contrastive lexical dataset $\mathbb{D}_\text{contrastive}$ \footnote{Sentiment lexicon used is 
from \url{https://www.cs.uic.edu/~liub/FBS/sentiment-analysis.html\#lexicon}.}. An evaluation protocol is designed to examine the effect of various corruption types.

\subsection{Segmentation Corruption under Noise}
A PLM consists of a tokenizer $\operatorname{Seg}(\cdot)$, which segments a given word $w$ into a sequence of subwords, i.e., $\operatorname{Seg}(w)=\left(\Tilde{w}_{1}, ..., \Tilde{w}_{K}\right)$, and a PLM encoder $\operatorname{Enc}(\cdot)$, which takes $\operatorname{Seg}(w)$ and outputs a word representation.
Formally, the segmentation of a canonical word $\operatorname{Seg}(w)$ can be represented as a set $\mathbb{S}$, while the segmentation of a noisy word $\operatorname{Seg}(\tilde{w})$ can be represented as set $\mathbb{\Tilde{S}}=\{\Tilde{w}_1, ..., \Tilde{w}_K\}$. We can then utilize set operations to define the overlap set (consisting of retained subwords), the missing set, and the additive set (comprising additional tokens that are not present in $\mathbb{S}$) as $\mathbb{O} = \mathbb{S} \cap \mathbb{\Tilde{S}}$, $\mathbb{M}=\mathbb{S}-\mathbb{O}$ and $\mathbb{A} = \mathbb{\Tilde{S}}-\mathbb{O}$, respectively. 

The set data structure cannot count duplicated tokens, which frequently occur in additive corruption scenarios, such as the additive (affix) corruption example presented in Table \ref{tab:subword_sets}. Hence, we utilize a multiset implementation of $\mathbb{S}$ and $\mathbb{\Tilde{S}}$ since such a data structure also stores the frequencies of elements, helping us assess the impact of duplicated tokens.
Since the multiset implementation only includes unique elements without considering their order of appearance, we further differentiate the two types of additive corruption by iteratively comparing elements from two queue implementations of $\operatorname{Seg}(w)$ and $\operatorname{Seg}(\Tilde{w})$.

In this study, we distinguish a unique category of corruption referred to as ``intact corruption'' from complete corruption, as the canonical words in this category (with whole-word vectors) remain unchanged. In total, there are six different types of corruption, as outlined in Table \ref{tab:subword_sets}. 

\paragraph{Identification of corruption types.}
During the evaluation, we need to filter each word pair according to its corruption type.
First, we segment each word pair in $\mathbb{D_\text{contrastive}}$ by a model-specific tokenizer $\operatorname{Seg}(\cdot)$ into subwords $(\mathbb{S}, \mathbb{\Tilde{S}})$.
We then identify the corruption type according to the following conditions: 
1) Complete corruption: $\mathbb{S}$ and $\mathbb{\Tilde{S}}$ are disjoint, i.e., $\mathbb{O}=\emptyset$. If the length of the missing set $\mathbb{M}$ is 1, this noise leads to intact corruption; 
2) Partial corruption: the corruption only occurs to one of the subwords (i.e., the one in $\mathbb{M}$), and the other subwords (i.e., those in $\mathbb{O}$) are not affected. 
The prerequisite is that there exist more than one subwords in the original segmentation set $\mathbb{S}$.
We can find such word pairs satisfy $\mathbb{M}, \mathbb{O}, \mathbb{A} \neq \emptyset$;
3) The conditions for additive corruption and missing corruption are $\mathbb{S} \in \mathbb{\Tilde{S}}$ (or $\mathbb{M} = \emptyset$) and $\mathbb{\Tilde{S}} \in \mathbb{S}$ (or $\mathbb{A} = \emptyset$), respectively. 
\footnote{See \url{https://github.com/xinzhel/word_corruption/blob/main/word_corruption.py} for concrete implementation.}

\subsection{Creation of Contrastive Dataset} \label{sec:contrastive_dataset}
Most prior noisy datasets added noise to sentences, not individual words \citep{belinkov2017synthetic,kumar-etal-2020-noisy,warstadt-etal-2019-neural,hagiwara-mita-2020-github}. Besides, as contrastive datasets containing both the original and noisy form of a word are not readily available, we create our own lexical dataset which includes both forms. Examples of the generated dataset can be found in Table \ref{tab:noisy_words}.

\begin{table*}[h!]
    \centering
    \begin{tabular}{l|l|l|l}
       Canonical Words & Keyboard & Swap  & Letter-reduplication \\
         \hline
         \hline
         bad 
            & NA
            & NA
            &  badddddddd, baaaadddd, bbbbaaaaddddd \\
        
        crazy
            & craxy
            & carzy
            & crazzyyyyyyyyy, crazzzzzy\\
            
        amazing 
            & amazijg
            & amzaing
            & amazing, amazinnng, amazinggg, amaaazzziiingggg \\
    \bottomrule
    \end{tabular}
    \caption{Examples of contrastive datasets with canonical-noisy word pairs. Three types of noise models are applied: Swap-typos, Keyboard typos and letter reduplication. NA: we discard generated noisy words since typos on these words generate noisy words that are even unrecognizable to humans. }
 
    \label{tab:noisy_words}
\end{table*}

\paragraph{Noise models.}

Two sources of noise models are used to generate the lexical dataset. Findings given in Appendix \ref{app:plm_by_noise} indicate that both types of noise models have comparable effects on model performance.

1) \textbf{Naturally and frequently occurring typos.}
Users often type neighboring keys due to mobile penetration across the globe and fat finger problem \citep{kumar-etal-2020-noisy}, while typing quickly may result in swapping two letters \citet{belinkov2017synthetic}. We refer to them as \textit{Keyboard} and \textit{Swap} typos, respectively. Our implementation of these typos is based on \citet{wang2021textflint}.
Specifically, for \textit{Keyboard}, we only use letters in the English alphabet within one keyboard distance as the substitute symbols. Further, we avoid unrecognizable word forms (e.g., ``bad$\rightarrow$bqd'' or ``top $\rightarrow$ tpp'') by selecting words with more than four characters.

According to the psycholinguistic study \citep{davis2003psycholinguistic}, to make noisy words recognizable for humans, we only apply noise to the middle characters and keep characters at the beginning and the end. 
Besides such a constraint, \textit{Swap} typo also requires at least two distinct characters in the middle for swapping. However, words like ``aggressive'' can still be transformed into the same word by swapping ``ss'', so we transform them until we get a distinct word. 
Finally, we set a one-edit constraint for typos.

2) \textbf{Non-standard orthography.} 
We gather words with letter reduplication from 1.6 million tweets \citep{go2009twitter}. To create the canonical and noisy word pairs, we match specific noisy word forms (e.g. words with repeated letters for emphasis) to their corresponding canonical forms (a sequence of definite characters). We use simple regular expression patterns to search for words with repeated letters \footnote{For example, pattern ``\textbackslash bb+a+d+'' for ``bad'' matches ``badddddddd''.}.
Examples in Table \ref{tab:noise_corrupt} show how effective these types of noise are in triggering different types of segmentation corruption. 

\begin{table*}[h!]
    \centering
    \small
    \begin{subtable}[!h]{0.5\textwidth}
    \centering
    \begin{tabular}{lrrr}
        \toprule
        Tokenizers & Intact & Complete &  \multicolumn{1}{c}{Partial}  \\
        &&&\\
        \midrule
        BERT   
        & 0.36
        & 0.14
        & 0.49
        \\
        RoBERTa   
        & 0.46
        & 0.12
        & 0.42
        \\
        ALBERT    
        & 0.38
        & 0.13
        & 0.49
        \\
        \bottomrule
    \end{tabular}
    \caption{Typos.}
    \end{subtable}
    \begin{subtable}[!h]{0.48\textwidth}
    \centering
    \begin{tabular}{rrrrr}
        \toprule
         Intact & Complete & Partial & \multicolumn{2}{c}{Additive} \\
         & & & affix & infix \\
        \midrule
        0.70
        & 0.02
        & 0.06
        & 0.22 
        & 0\\
        0.61
        & 0
        & 0.06
        & 0.30 
        & 0.01 \\
        0.61
        & 0.02
        & 0.06
        & 0.29 
        & 0.02 \\
        \bottomrule
    \end{tabular}
    \caption{Letter Reduplication.}
    \end{subtable}
    \caption{Frequency of each segmentation corruption.}
    
    \label{tab:noise_corrupt}
\end{table*}

\paragraph{Data-generating process.}
We create a contrastive dataset, $\mathbb{D_\text{contrastive}}$, by applying the noise models to the lexical dataset $\mathbb{D_\text{canonical}}$, which contains words in their canonical form.  
The noise models are applied to each word in $\mathbb{D_\text{canonical}}$ to create two misspelled words. Additionally, a random number of noisy words is extracted from the collection of 1.6 million tweets. As for the lexical dataset, $\mathbb{D_\text{canonical}}$, we choose adjectives from a sentiment lexicon that, by definition, provides positive or negative sentiment labels for use with downstream classifiers.

\paragraph{Evaluation.}
To assess the extent to which the meanings of noisy words diverge from the standard word forms, we calculate the cosine similarity between $\operatorname{Enc}(\mathbb{S})$ and $\operatorname{Enc}(\mathbb{\Tilde{S}})$.
For words that consist of multiple subwords, we aggregate their vectors into a single representation by averaging the token embeddings obtained from the PLMs.
It is important to note that the output embedding spaces of PLMs exhibit varying levels of anisotropy \citep{ethayarajh-2019-contextual, yan-etal-2021-consert, gao2018representation}. Thus, the similarity scores cannot be directly compared across different models. 
It is necessary to set a baseline by computing the similarity between $\operatorname{Enc}(\mathbb{S})$ and a random embedding (we use the embedding of token ``the'', i.e., $\operatorname{Enc}(\text{the})$).

Additionally, we fine-tune downstream classifiers denoted as  $y=\operatorname{Cls}(x)$, where $y$ represents an arbitrary semantic dimension and $x$ corresponds to the encoded representation obtained from the PLMs $\operatorname{Enc}(\operatorname{Seg}(\cdot))$. 
We focus on sentiment classification as individuals frequently use sentiment words creatively on social media to express their emotions. To conduct our experiments, the sentiment of each word and its noisy variations is derived from the sentiment lexicon.

To gauge the semantic deviation caused by noise, we measure the accuracy of the noisy counterparts of words that are accurately classified in their original form.

\section{Experimental Results} 
Experiments are performed on three widely used PLMs: \textbf{BERT\textsubscript{BASE}}, \textbf{RoBERTa\textsubscript{BASE}} and \textbf{ALBERT\textsubscript{BASE}} (See Appendix \ref{app:plms} for details).
\textbf{BERT} \citep{devlin-etal-2019-bert} accepts inputs from a Wordpiece tokenizer \citep{schuster2012wordpiece}, while \textbf{RoBERTa} \citep{liu2020roberta}, another popular frequent-based segmentation scheme, uses BPE \citep{sennrich-etal-2016-neural}. For comparison, we include \textbf{ALBERT} \citep{lan2020albert} with a probabilistic tokenizer called Sentencepiece \citep{kudo-richardson-2018-sentencepiece}.

\begin{table*}[h!]
    \centering
    \begin{subtable}[!h]{0.6\textwidth}
    \centering
    \small
    \begin{tabular}{lrrrrr}
        \toprule
        Models & Intact & Complete & Partial & Additive & Baseline  \\
        \midrule
        BERT   
        & 0.29
        & 0.41
        & \cellcolor{black!15}0.58
        & \cellcolor{black!25}0.69
        & 0.69 \\
        RoBERTa   
        & 0.54
        & 0.66
        & \cellcolor{black!15}0.76
        & \cellcolor{black!25}0.85
        & 0.72 \\
        ALBERT    
        & 0.41
        & 0.47
        & \cellcolor{black!15}0.62
        & \cellcolor{black!25}0.74
        & 0.68\\
        
        \bottomrule
    \end{tabular}
    \caption{Similarity.}
    \end{subtable}
    \small
    \begin{subtable}[!h]{0.33\textwidth}
    \centering
    \begin{tabular}{rrrrr}
        \toprule
         Intact & Complete & Partial & Additive\\
        \midrule
        0.56
        & 0.65
        & \cellcolor{black!15}0.8
        & \cellcolor{black!25}0.91\\
        0.66
        & 0.60
        &\cellcolor{black!15}0.75 
        &\cellcolor{black!25}0.95\\
        0.61
        & 0.63
        & \cellcolor{black!15}0.76
        & \cellcolor{black!25}0.93 \\
        \bottomrule
    \end{tabular}
    \caption{Accuracy. }
    \end{subtable}
    \caption{Performance of PLMs under various types of corruption. Similarity scores of pretrained representations and accuracy of downstream classifiers are evaluated. The best result per row is highlighted in gray, and the second-best is in light gray. As a baseline, we compare the similarity scores between canonical and random words (``the'' used). The unaffected accuracy is 1 since the canonical forms selected for evaluation are always correctly predicted.}
    \label{tab:plm_under_corrupt1}    
\end{table*}

\begin{table}[h!]
    \centering
    \begin{subtable}[!h]{0.25\textwidth}
    \centering
    \small
    \begin{tabular}{lrr}
        \toprule
        Models & Infix & Suffix  \\
        \midrule
        BERT   
        & 0.59
        & \cellcolor{black!25}0.70 \\
        RoBERTa   
        & 0.85 
        & \cellcolor{black!25}0.95 \\
        ALBERT   
        & 0.66
        & \cellcolor{black!25}0.74 \\
       
        \bottomrule
    \end{tabular}
    \caption{Similarity.}
    \end{subtable}
    \small
    \begin{subtable}[!h]{0.2\textwidth}
    \centering
    \begin{tabular}{rr}
        \toprule
         Infix & Suffix\\
        \midrule
        0.74
        & \cellcolor{black!25}0.91 \\
        0.95
        & \cellcolor{black!25}1 \\
        0.82
        & \cellcolor{black!25}0.94 \\
        \bottomrule
    \end{tabular}
    \caption{Accuracy. }
    \end{subtable}
    \caption{Comparison of two types of additive corruption. }
    \label{tab:types_additive}    
\end{table}

\paragraph{Subwords retention is important for maintaining the correct semantics.}
Table~\ref{tab:plm_under_corrupt1} shows the severity of semantic deviation for each type of corruption. 
Generally, the more subwords the segmentation retains, the better the semantics are maintained (additive corruption $>$ partial corruption $>$ complete and intact corruption).
Under additive corruption, the PLMs can always maintain more semantics from noisy words than random words (the baseline), while only RoBERTa has similarity score higher than the baseline under partial corruption. 
All the PLMs cannot infer word meaning from complete corruption. 

What subwords, if retained, would enhance the comprehension of PLMs? We find that partial corruption can preserve word meaning if it retains a significant portion of the words, such as ``upset'' for ``upsetting'' or ``phenomena'' for ``phenomenal'' (See Appendix \ref{app:good_bad_cases}). This is backed up by the finding that PLMs have the capability of learning morphological information, where stems contain more semantic meaning in a word compared to smaller components such as inflectional morphemes \citep{hofmann-etal-2021-superbizarre}.

\paragraph{Are words more impacted by noise under complete corruption if their meaning is stored in the embeddings?}
According to \citet{hofmann-etal-2021-superbizarre}, if a word is represented as a single vector, PLMs can access its meaning directly from the embedding (referred to as the ``storage route'') instead of deducing it from the combination of subwords (known as the ``computation route''). 
We presume that PLMs struggle to maintain the original meaning of these words when exposed to noise. We classify this type of corruption as ``intact corruption'', which is a particular variation to complete corruption.
To validate our assumption, we evaluate the performance of PLMs on words under intact corruption. Results show that words with intact corruption consistently perform worse than those with complete corruption, despite both having completely distinct subwords. 
Although intact corruption consistently yields the lowest similarity score, the PLMs may still be able to better infer some semantic dimensions, such as sentiment, under intact corruption compared to complete corruption. (Appendix \ref{app:corruption_under_diff_noise}).

\paragraph{Presence of additive subwords can damage the meaning of words, particularly when they are inserted in the middle of other subwords.}
In some cases, words under additive corruption (keeping all subwords) can perform worse than those under partial corruption (keeping only some subwords), as seen in the letter reduplication experiment (Appendix \ref{app:corruption_under_diff_noise}). The finding suggests that the retention of subwords is not the only factor impacting the performance of PLMs. To uncover other factors affecting the word meaning, we analyzed 10 worst and best instances for each corruption type based on similarity scores (Appendix \ref{app:good_bad_cases}). All the poorly performing cases have incorrect predictions, further highlighting the damaging impact on semantic meaning.
The results show that the number of additive tokens (i.e., the cardinality of $\mathbb{A}$) is a distinct feature between good and bad instances. All the good cases have only 1 additive token, while the bad cases have at least 2 additive tokens (3.8 for partial corruption and 8.7 for additive corruption on average). 

\begin{figure}[h!]
    \centering
    \small
    \includegraphics[width=0.4\textwidth]{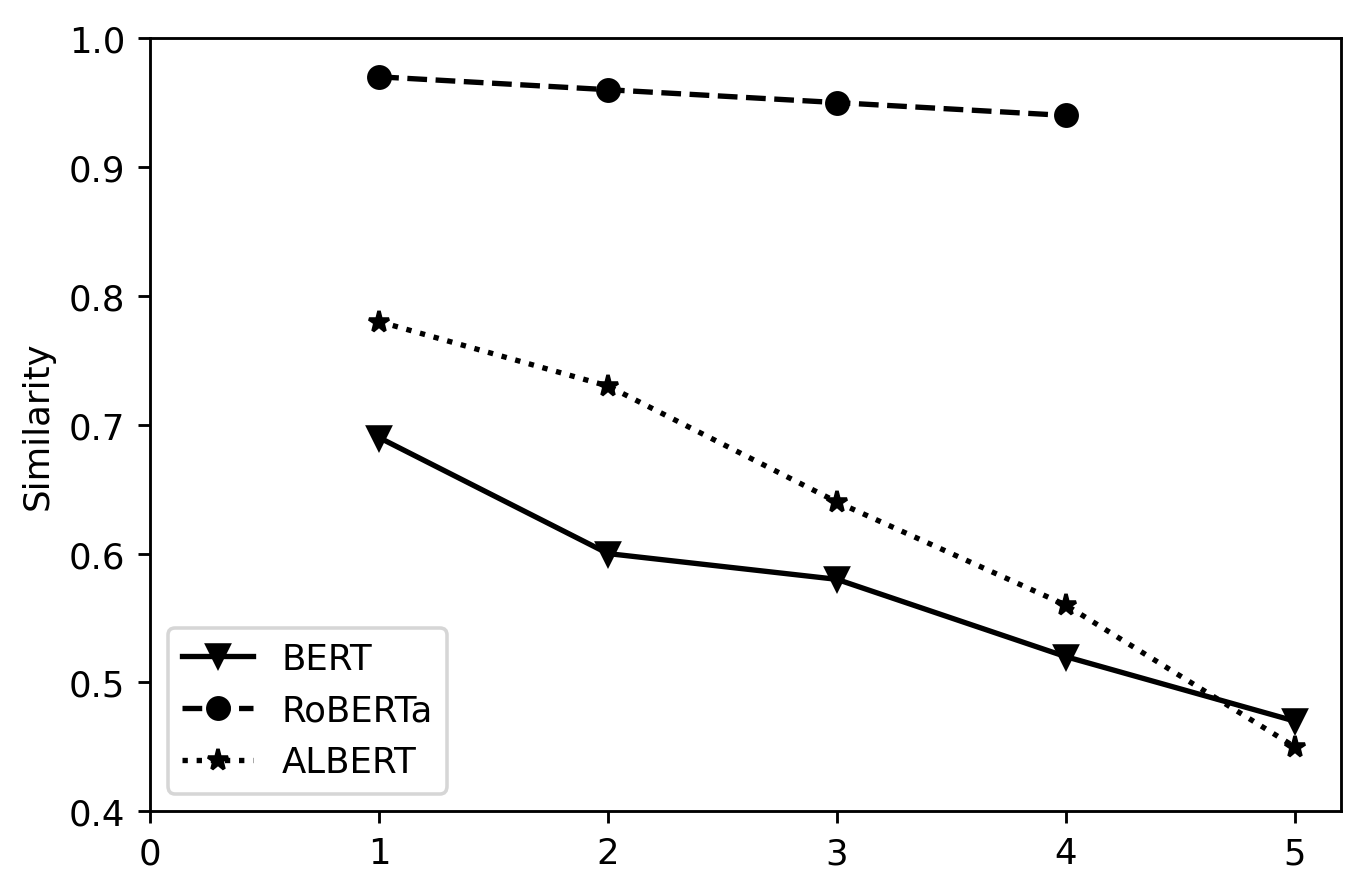}
    \caption{Correlation between the number of additive subwords and the cosine similarity of noisy words with their canonical forms. The range of quantity of additive subwords is subject to change depending on the tokenizer used.}
    \label{fig:lexicon_m_score_vs_sim}
\end{figure}

Thus, our hypothesis is that as the number of additive subwords increases, PLMs will have difficulty determining the correct meaning of words. We test the hypothesis by examining the performance of PLMs on both additive and intact corruption, where the missing and overlap sets remain constant.
For additive corruption, we limit our experiments to only one unique additive subword and vary its frequency. We find 23 words with at least 3 noisy versions, each creating an additive set with the same element but different multiplicities.
Take ``amazing'' as an example: one of its noisy instances (``amazinggggggg'') has the multiplicity of 3 according to its additive set $\mathbb{A}=\{\text{``gg'', ``gg'', ``gg''}\}$ while ``gg'' only appears twice in another instance (``amazinggggg''). 
We sort every collection of noisy words in either of two ways, depending on the similarity scores or the multiplicities of additive tokens.  In 17 out of 23 collections, these two sorting criteria produce identical results. This discovery also holds true for intact corruption, where the subwords within an additive set are typically diverse.
Figure \ref{fig:lexicon_m_score_vs_sim} illustrates a strong negative correlation between the number of additive tokens and the average similarity of noisy words for all the three models under intact corruption, where the sizes of missing sets and overlap sets are fixed to 1 and 0.

Besides, as shown in Table~\ref{tab:types_additive}, additive subwords placed within subwords cause more harm than those that act as suffixes.



\section{Conclusion}
We proposed the CoLeS framework which can evaluate how corrupt segmentation under noise affects PLMs' understanding.
The experimental results show that three challenges can impair the PLMs' understanding of noisy words: insertion of additive subwords (especially within existing subwords), loss of original subwords, and incapacity of computing the word meanings through the aggregation of smaller subword units.

\paragraph{Reproducibility.}
Data and source code for noisy data generation, corruption types identification and PLMs' performance evaluation are released on Github \footnote{\url{https://github.com/xinzhel/word_corruption}}.

\section*{Limitations}
The omission of missing corruption from the evaluation process is justified due to its infrequent occurrence in real-world scenarios (refer to Appendix \ref{app:missing} for elaboration). Nevertheless, further investigation into rare instances of missing corruption may be warranted for research purposes. Our evaluation of language models was limited to auto-encoders based on the BERT architecture. Future studies are anticipated to expand the scope of PLMs under consideration \footnote{
Instructions within our codebase facilitate the evaluation of various types of pre-trained language models accessible via Huggingface\url{https://huggingface.co/models}.}. 

\bibliography{anthology,custom}

\begin{thebibliography}{26}
\expandafter\ifx\csname natexlab\endcsname\relax\def\natexlab#1{#1}\fi

\bibitem[{Baron(2015)}]{baron2015words}
Naomi~S Baron. 2015.
\newblock \emph{Words onscreen: The fate of reading in a digital world}.
\newblock Oxford University Press, USA.

\bibitem[{Belinkov and Bisk(2018)}]{belinkov2017synthetic}
Yonatan Belinkov and Yonatan Bisk. 2018.
\newblock \href {https://openreview.net/forum?id=BJ8vJebC-} {Synthetic and
  natural noise both break neural machine translation}.
\newblock In \emph{ICLR}.

\bibitem[{Davis(2003)}]{davis2003psycholinguistic}
Matt Davis. 2003.
\newblock Psycholinguistic evidence on scrambled letters in reading.

\bibitem[{Devlin et~al.(2019)Devlin, Chang, Lee, and
  Toutanova}]{devlin-etal-2019-bert}
Jacob Devlin, Ming-Wei Chang, Kenton Lee, and Kristina Toutanova. 2019.
\newblock \href {https://doi.org/10.18653/v1/N19-1423} {{BERT}: Pre-training of
  deep bidirectional transformers for language understanding}.
\newblock In \emph{NAACL}.

\bibitem[{El~Boukkouri et~al.(2020)El~Boukkouri, Ferret, Lavergne, Noji,
  Zweigenbaum, and Tsujii}]{el-boukkouri-etal-2020-characterbert}
Hicham El~Boukkouri, Olivier Ferret, Thomas Lavergne, Hiroshi Noji, Pierre
  Zweigenbaum, and Jun{'}ichi Tsujii. 2020.
\newblock \href {https://doi.org/10.18653/v1/2020.coling-main.609}
  {{C}haracter{BERT}: Reconciling {ELM}o and {BERT} for word-level
  open-vocabulary representations from characters}.
\newblock In \emph{CoLing}, pages 6903--6915.

\bibitem[{Ethayarajh(2019)}]{ethayarajh-2019-contextual}
Kawin Ethayarajh. 2019.
\newblock \href {https://doi.org/10.18653/v1/D19-1006} {How contextual are
  contextualized word representations? {C}omparing the geometry of {BERT},
  {ELM}o, and {GPT}-2 embeddings}.
\newblock In \emph{EMNLP-IJCNLP}, pages 55--65, Hong Kong, China.

\bibitem[{Gao et~al.(2019)Gao, He, Tan, Qin, Wang, and
  Liu}]{gao2018representation}
Jun Gao, Di~He, Xu~Tan, Tao Qin, Liwei Wang, and Tieyan Liu. 2019.
\newblock \href {https://openreview.net/forum?id=SkEYojRqtm} {Representation
  degeneration problem in training natural language generation models}.
\newblock In \emph{ICLR}.

\bibitem[{Go et~al.(2009)Go, Bhayani, and Huang}]{go2009twitter}
Alec Go, Richa Bhayani, and Lei Huang. 2009.
\newblock \href
  {https://www-cs.stanford.edu/people/alecmgo/papers/TwitterDistantSupervision09.pdf}
  {Twitter sentiment classification using distant supervision}.
\newblock \emph{CS224N project report, Stanford}.

\bibitem[{Hagiwara and Mita(2020)}]{hagiwara-mita-2020-github}
Masato Hagiwara and Masato Mita. 2020.
\newblock \href {https://aclanthology.org/2020.lrec-1.835} {{G}it{H}ub typo
  corpus: A large-scale multilingual dataset of misspellings and grammatical
  errors}.
\newblock In \emph{Language Resources and Evaluation Conference (LREC)}.

\bibitem[{Heath(2018)}]{heath2018orthography}
Maria Heath. 2018.
\newblock Orthography in social media: Pragmatic and prosodic interpretations
  of caps lock.
\newblock \emph{Proceedings of the Linguistic Society of America}.

\bibitem[{Hofmann et~al.(2021)Hofmann, Pierrehumbert, and
  Sch{\"u}tze}]{hofmann-etal-2021-superbizarre}
Valentin Hofmann, Janet Pierrehumbert, and Hinrich Sch{\"u}tze. 2021.
\newblock \href {https://doi.org/10.18653/v1/2021.acl-long.279} {Superbizarre
  is not superb: Derivational morphology improves {BERT}{'}s interpretation of
  complex words}.
\newblock In \emph{ACL-IJCNLP}.

\bibitem[{Kudo and Richardson(2018)}]{kudo-richardson-2018-sentencepiece}
Taku Kudo and John Richardson. 2018.
\newblock \href {https://doi.org/10.18653/v1/D18-2012} {{S}entence{P}iece: A
  simple and language independent subword tokenizer and detokenizer for neural
  text processing}.
\newblock In \emph{EMNLP: System Demonstrations}, pages 66--71.

\bibitem[{Kumar et~al.(2020)Kumar, Makhija, and Gupta}]{kumar-etal-2020-noisy}
Ankit Kumar, Piyush Makhija, and Anuj Gupta. 2020.
\newblock \href {https://doi.org/10.18653/v1/2020.wnut-1.3} {Noisy text data:
  Achilles{'} heel of {BERT}}.
\newblock In \emph{Proceedings of the Sixth Workshop on Noisy User-generated
  Text (W-NUT 2020)}.

\bibitem[{Lan et~al.(2020)Lan, Chen, Goodman, Gimpel, Sharma, and
  Soricut}]{lan2020albert}
Zhenzhong Lan, Mingda Chen, Sebastian Goodman, Kevin Gimpel, Piyush Sharma, and
  Radu Soricut. 2020.
\newblock \href {https://openreview.net/forum?id=H1eA7AEtvS} {{ALBERT:} {A}
  lite {BERT} for self-supervised learning of language representations}.
\newblock In \emph{ICLR}.

\bibitem[{Liu et~al.(2022)Liu, Yu, Hu, Li, Lin, Ma, Yang, and
  Wen}]{liu-etal-2022-character}
Aiwei Liu, Honghai Yu, Xuming Hu, Shu{'}ang Li, Li~Lin, Fukun Ma, Yawen Yang,
  and Lijie Wen. 2022.
\newblock \href {https://aclanthology.org/2022.emnlp-main.522} {Character-level
  white-box adversarial attacks against transformers via attachable subwords
  substitution}.
\newblock In \emph{EMNLP}.

\bibitem[{Liu et~al.(2019)Liu, Ott, Goyal, Du, Joshi, Chen, Levy, Lewis,
  Zettlemoyer, and Stoyanov}]{liu2020roberta}
Yinhan Liu, Myle Ott, Naman Goyal, Jingfei Du, Mandar Joshi, Danqi Chen, Omer
  Levy, Mike Lewis, Luke Zettlemoyer, and Veselin Stoyanov. 2019.
\newblock Roberta: A robustly optimized bert pretraining approach.
\newblock \emph{arXiv preprint arXiv:1907.11692}.

\bibitem[{Rawlinson(2007)}]{rawlinson2007letter}
Graham Rawlinson. 2007.
\newblock \href {https://doi.org/10.1109/MAES.2007.327521} {The significance of
  letter position in word recognition}.
\newblock \emph{IEEE Aerospace and Electronic Systems Magazine}.

\bibitem[{Ritter et~al.(2010)Ritter, Cherry, and
  Dolan}]{ritter-etal-2010-unsupervised}
Alan Ritter, Colin Cherry, and Bill Dolan. 2010.
\newblock \href {https://aclanthology.org/N10-1020} {Unsupervised modeling of
  {T}witter conversations}.
\newblock In \emph{NAACL-HLT}, pages 172--180.

\bibitem[{Schuster and Nakajima(2012)}]{schuster2012wordpiece}
Mike Schuster and Kaisuke Nakajima. 2012.
\newblock \href {https://doi.org/10.1109/ICASSP.2012.6289079} {Japanese and
  korean voice search}.
\newblock In \emph{2012 IEEE International Conference on Acoustics, Speech and
  Signal Processing (ICASSP)}, pages 5149--5152.

\bibitem[{Sennrich et~al.(2016)Sennrich, Haddow, and
  Birch}]{sennrich-etal-2016-neural}
Rico Sennrich, Barry Haddow, and Alexandra Birch. 2016.
\newblock \href {https://doi.org/10.18653/v1/P16-1162} {Neural machine
  translation of rare words with subword units}.
\newblock In \emph{ACL}.

\bibitem[{Vaswani et~al.(2017)Vaswani, Shazeer, Parmar, Uszkoreit, Jones,
  Gomez, Kaiser, and Polosukhin}]{vaswani2017attention}
Ashish Vaswani, Noam Shazeer, Niki Parmar, Jakob Uszkoreit, Llion Jones,
  Aidan~N Gomez, \L~ukasz Kaiser, and Illia Polosukhin. 2017.
\newblock \href
  {https://proceedings.neurips.cc/paper/2017/file/3f5ee243547dee91fbd053c1c4a845aa-Paper.pdf}
  {Attention is all you need}.
\newblock In \emph{Advances in Neural Information Processing Systems}.

\bibitem[{Wang et~al.(2021)Wang, Liu, Gui, Zhang et~al.}]{wang2021textflint}
Xiao Wang, Qin Liu, Tao Gui, Qi~Zhang, et~al. 2021.
\newblock \href {https://doi.org/10.18653/v1/2021.acl-demo.41} {Textflint:
  Unified multilingual robustness evaluation toolkit for natural language
  processing}.
\newblock In \emph{ACL-IJCNLP: System Demonstrations}, Online.

\bibitem[{Warstadt et~al.(2019)Warstadt, Singh, and
  Bowman}]{warstadt-etal-2019-neural}
Alex Warstadt, Amanpreet Singh, and Samuel~R. Bowman. 2019.
\newblock \href {https://doi.org/10.1162/tacl_a_00290} {Neural network
  acceptability judgments}.
\newblock \emph{TACL}.

\bibitem[{Wolf et~al.(2020)Wolf, Debut, Sanh, Chaumond, Delangue, Moi, Cistac,
  Rault, Louf, Funtowicz, Davison, Shleifer, von Platen, Ma, Jernite, Plu, Xu,
  Le~Scao, Gugger, Drame, Lhoest, and Rush}]{wolf-etal-2020-transformers}
Thomas Wolf, Lysandre Debut, Victor Sanh, Julien Chaumond, Clement Delangue,
  Anthony Moi, Pierric Cistac, Tim Rault, Remi Louf, Morgan Funtowicz, Joe
  Davison, Sam Shleifer, Patrick von Platen, Clara Ma, Yacine Jernite, Julien
  Plu, Canwen Xu, Teven Le~Scao, Sylvain Gugger, Mariama Drame, Quentin Lhoest,
  and Alexander Rush. 2020.
\newblock \href {https://doi.org/10.18653/v1/2020.emnlp-demos.6} {Transformers:
  State-of-the-art natural language processing}.
\newblock In \emph{EMNLP: System Demonstrations}.

\bibitem[{Yan et~al.(2021)Yan, Li, Wang, Zhang, Wu, and
  Xu}]{yan-etal-2021-consert}
Yuanmeng Yan, Rumei Li, Sirui Wang, Fuzheng Zhang, Wei Wu, and Weiran Xu. 2021.
\newblock \href {https://doi.org/10.18653/v1/2021.acl-long.393} {{C}on{SERT}: A
  contrastive framework for self-supervised sentence representation transfer}.
\newblock In \emph{ACL-IJCNLP}.

\bibitem[{Zhuang and Zuccon(2022)}]{zhuang2022typos}
Shengyao Zhuang and Guido Zuccon. 2022.
\newblock \href {https://doi.org/10.1145/3477495.3531951} {Characterbert and
  self-teaching for improving the robustness of dense retrievers on queries
  with typos}.
\newblock SIGIR.

\end{thebibliography}
\bibliographystyle{acl_natbib}
\clearpage
\onecolumn
\appendix

\section{Fine-tuning Pretrained Language Models} \label{app:plms}
All the PLMs use BERT-based architecture, i.e., the encoding part of the transformer \citep{vaswani2017attention}.
\textbf{BERT\textsubscript{BASE}} (110M parameters) and \textbf{RoBERTa\textsubscript{BASE}} (125M parameters) are pretrained on BookCorpus and Wikipedia as masked language models. Only the pretraining of \textbf{ALBERT\textsubscript{BASE}} (11M parameters) includes extra news and web data \citep{wolf-etal-2020-transformers}.
They are then fine-tuned for sentiment classification on the SST-2 dataset. All the models are publicly available on the Huggingface Hub website \url{https://huggingface.co/textattack}.  Some configurations are shown as below. 
The BERT and RoBERTa models are fine-tuned using
a learning rate of $2 e^{-5}$ with no scheduling employed. The batch size is set to 32, and the training process spans 3 epochs, maintaining a gradient norm of 1.  
ALBERT is fined-tuned with a learning rate of $3 e^{-5}$, a batch size of 32, and a total of 5 training epochs.

\section{Good and Bad Cases} \label{app:good_bad_cases}
\begin{figure}[h!]
    \centering
    \includegraphics[width=\textwidth]{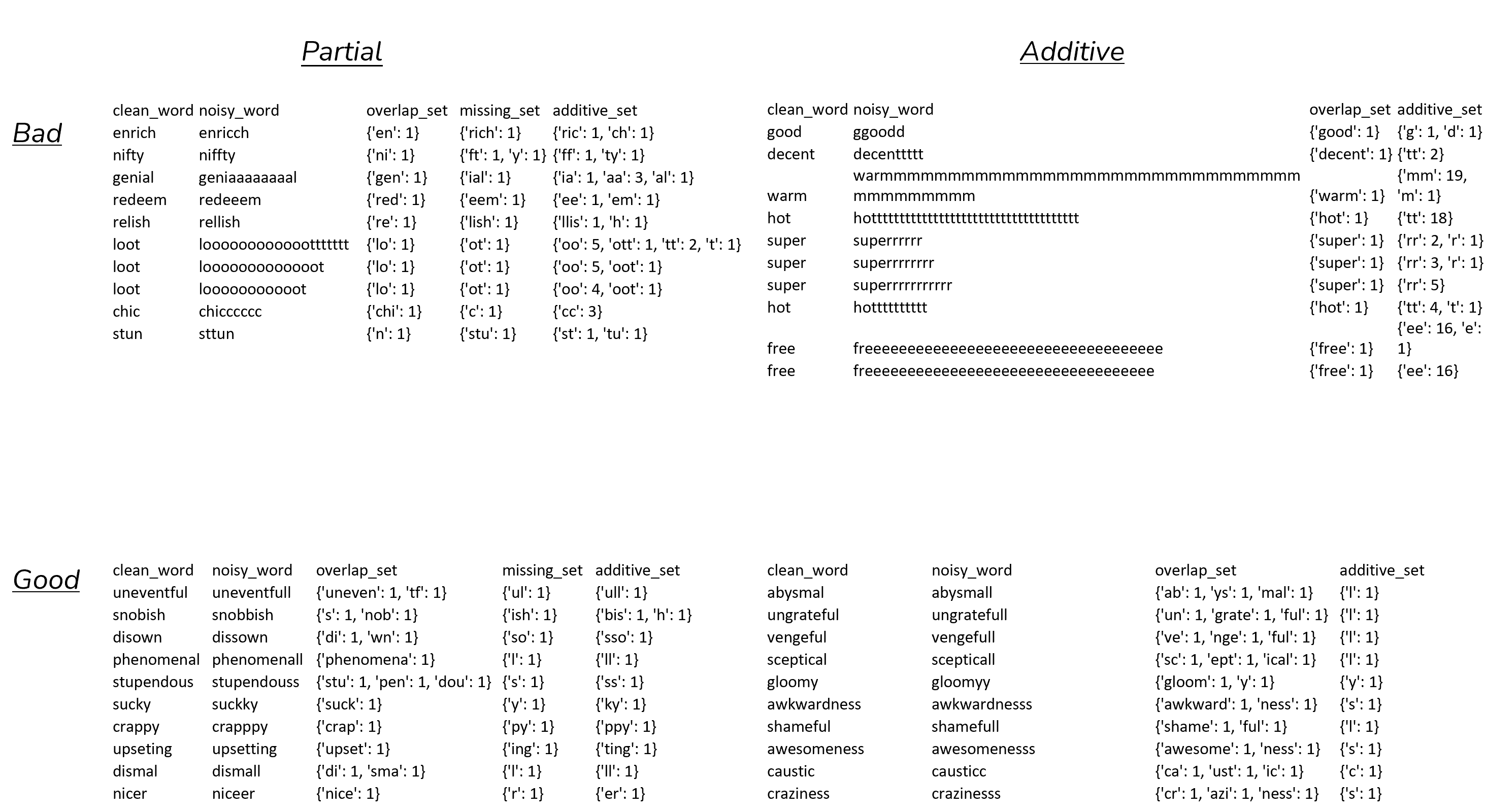}
    \caption{Good and bad cases of partial and additive corruption under letter reduplication.}
    \label{fig:good_bad_cases}
\end{figure}

Figure \ref{fig:good_bad_cases} shows the good and bad cases of partial and additive corruption under letter reduplication.

\section{PLM Robustness to Segmentation Corruption under Different Types of Noise} \label{app:corruption_under_diff_noise}

Table \ref{tab:plm_under_corrupt2} displays the robustness of PLMs to segmentation corruption under various forms of noise. The results are largely consistent with those seen in Table \ref{tab:plm_under_corrupt1}. However, we notice that for letter reduplication, PLMs may perform worse with additive corruption than with partial corruption.  Additionally, the accuracy of intact corruption can be better than that of complete corruption, despite they consistently having the lowest similarity score.

\begin{table*}[h!]
    
    \centering
    \begin{subtable}[!h]{0.5\textwidth}
    \centering
    \begin{tabular}{lrrrr}
        \toprule
        Models & Intact & Complete & Partial & Additive \\
        \midrule
        BERT   
        & 0.24
        & 0.34
        & \cellcolor{black!15}0.62 
        & \cellcolor{black!25}0.69 \\
        RoBERTa   
        & 0.54
        & 0.73
        & \cellcolor{black!15}0.8
        & \cellcolor{black!25}0.85 \\
        ALBERT    
        & 0.42
        & 0.53
        & \cellcolor{black!25}0.77
        & \cellcolor{black!15}0.74 \\
        \bottomrule
    \end{tabular}
    \caption{Similarity.}
    \end{subtable}
    \begin{subtable}[!h]{0.48\textwidth}
    \centering
    \begin{tabular}{rrrr}
        \toprule
         Intact & Complete & Partial & Additive \\
        \midrule
        0.54
        & 0.47 
        & \cellcolor{black!25}0.92
        & \cellcolor{black!15}0.91\\
        0.54
        & 0.73
        &\cellcolor{black!15}0.87 &\cellcolor{black!25}0.95\\
        0.63
        & 0.79
        & \cellcolor{black!15}0.91
        & \cellcolor{black!25}0.93 \\
        \bottomrule
    \end{tabular}
    \caption{Accuracy.}
    \end{subtable}
    \caption*{Letter Reduplication}
    
    \bigskip
    \begin{subtable}[!h]{0.5\textwidth}
    \centering
    \begin{tabular}{lrrrr}
        \toprule
        Models & Intact & Complete & Partial & Additive \\
        \midrule
        BERT   
        & 0.34 
        & \cellcolor{black!15}0.41 
        & \cellcolor{black!25}0.58
        & / \\
        RoBERTa   
        & 0.55  
        & \cellcolor{black!15}0.65
        & \cellcolor{black!25} 0.75
        & /\\
        ALBERT    
        & 0.4
        & \cellcolor{black!15}0.47
        & \cellcolor{black!25}0.61
        & / \\
        \bottomrule
    \end{tabular}
    \caption{Similarity.}
    \end{subtable}
    \begin{subtable}[!h]{0.48\textwidth}
    \centering
    \begin{tabular}{rrrr}
        \toprule
         Intact & Complete & Partial & Additive \\
        \midrule
        0.59
        & \cellcolor{black!15}0.66
        & \cellcolor{black!25} 0.79
        & / \\
        \cellcolor{black!15}0.6
        &0.57
        &\cellcolor{black!25}0.74 
        &/ \\
        0.59
        & \cellcolor{black!15}0.61
        & \cellcolor{black!25}0.75
        & / \\
        \bottomrule
    \end{tabular}
    \caption{Accuracy.}
    \end{subtable}
    \caption*{Typos}
    \caption{Performance of PLMs under different types of corruption. \textit{Similarity scores of pretrained representations} and \textit{accuracy of downstream classifiers} are measured. The best result per row is highlighted
    in gray, the second-best is in light gray. There is no result for additive corruption under typos because intra-word noise (modifying characters except for the first and last characters) (i.e., typos) never results in additive corruption.
    Baseline similarity scores are calculated between canonical words and the word ``the''.}
    \label{tab:plm_under_corrupt2}
\end{table*}

\section{Noisy words for Missing Corruption} \label{app:missing}
As per the findings of Heath et al. \citep{heath2018orthography}, English word recognition by humans is predominantly influenced by consonants. Consequently, our investigation aims to identify abbreviations that disregard vowels and certain consonants when examining tweets. 
To be precise, an abbreviation is considered acceptable if its first letter and arbitrary consonants appear in a sequence that adheres to canonical words. 
For instance, the pattern of regular expression for term ``sorry'' can be ``\textbackslash bsr?r?y?'' in such cases. However, we find that even humans have difficulties in recognizing all these abbreviations. While the inclusion of all consonants may enhance human recognition, we contend that assessing this form of corruption is superfluous. This assertion stems from our demonstration in Table \ref{tab:missing} that such aggressive search criteria are improbable to produce missing corruption.

\begin{table}[h!]
    \centering
    \small
    \begin{tabular}{lr}
        \toprule
        Models & Intra   \\
        \midrule
        BERT   
        & 0.50\% \\
        RoBERTa   
        & 0.52\%  \\
        ALBERT   
        & 0.43\% \\
       
        \bottomrule
    \end{tabular}

    \caption{Proportion of abbreviations causing missing corruption. }
    \label{tab:missing}    
\end{table}

Provided below is a comprehensive inventory of the canonical words and their corresponding noisy counterparts responsible for inducing missing corruption. It is worth noting that these noisy words are completely imperceptible to human cognition.
\begin{itemize}
    \item 24 word pairs under RoBERTa:
enthral-enth,
upgradable-upgr,
abysmal-abys,
chintzy-chzy,
emphatic-emph,
enslave-ensl,
extraneous-extr,
implacable-impl,
implausible-impl,
implicate-impl,
imprudent-impr,
inflame-infl,
instable-inst,
intransigent-intr,
irksomeness-irks,
obscenity-obsc,
obtrusive-obtr,
ungrateful-ungr,
unscrupulous-unsc,
unsteadily-unst,
unsteadiness-unst,
unsteady-unst,
unsteady-unsty,
untruthful-untr;

    \item 
23 word pairs under BERT:
enthral-enth,
exemplar-expl,
exemplar-empl,
idyllic-idyl,
stylish-styl,
abysmal-abys,
brutish-brsh,
crummy-crmy,
enslave-ensl,
hysteric-hyst,
impenitent-impt,
incognizant-inct,
inconstant-inct,
inexplainable-inpl,
infamy-inmy,
inflame-infl,
irksomeness-irks,
obscenity-obsc,
obtrusive-obtr,
unscrupulous-unsc,
unspeakable-unsp,
untrue-untr,
untruthful-untr;

    \item 2 word pairs under ALBERT: enthral-enth, exemplar-exmp.

\end{itemize}

\section{Performance of PLMs under Different Noise} \label{app:plm_by_noise}
We compare the effect of two noise models ``Naturally and frequently occurring typos''
and ``Non-standard orthography'' with both the lexicon dataset and two sentential datasets.
For a fair comparison, we constrain the length of \textit{letter-reduplication} to 1. The accuracy of the noisy data and their standard deviation are reported in Table \ref{tab:plm_by_noise} and Table \ref{tab:std_by_noise}, respectively.
It can be seen that the types of noise models in our experiments have no much distinction on model performance, except for the \textit{Swap}. 

\begin{table*}[h!]
    \centering
    \begin{tabular}{llrrr}
        \toprule
       Data & Noise Type & BERT & RoBERTa & ALBERT\\
        \midrule
        \textbf{Accuracy} \\
        \multirow{5}{*}{SST-2}
        &  Clean            & 0.93 & 0.85  & 0.92 \\
        &  Keyboard         & 0.66 & 0.66  & 0.67 \\
        &  Swap        & 0.71 & 0.72  & 0.72 \\
        &  Letter-repetition & 0.63 & 0.7   & 0.65 \\
        \midrule
        \multirow{4}{*}{AG-News}
        &  Clean            & 0.95 & 0.8   & 0.92 \\
        &  Keyboard         & 0.88 & 0.62  & 0.86 \\
        &  Swap        & 0.89 & 0.62  & 0.86 \\
        &  Letter-repetition & 0.88 & 0.61  & 0.86 \\
        \bottomrule
        \textbf{Similarity} \\
        \multirow{3}{*}{Setiment Lexicon}
        &  Keyboard         & 0.39 & 1  & 0.47  \\
        &  Swap        & 0.45 & 1  & 0.5 \\
        &  Letter-repetition & 0.36 & 1  & 0.49 \\
        \midrule
        \multirow{5}{*}{SST-2}
        &  Keyboard         & 0.5  & 0.52  & 0.61 \\
        &  Swap        & 0.61 & 0.56  & 0.66 \\
        &  Letter-repetition & 0.46 & 0.55  & 0.58 \\
        
        \midrule
        \multirow{4}{*}{AG-News}
        &  Keyboard         & 0.85 & 0.47  & 0.72 \\
        &  Swap        & 0.87 & 0.5   & 0.75 \\
        &  Letter-repetition & 0.85 & 0.48  & 0.74 \\
        \bottomrule
    \end{tabular}
 
    \caption{Performance of PLMs under Different Noise}
    \label{tab:plm_by_noise}
\end{table*}

\begin{table}[h!]
    \centering
    \begin{tabular}{p{2cm}rrr}
        \toprule
       Data & BERT & RoBERTa & ALBERT\\
        \midrule
        \textbf{Similarity} && \\
        Lexicon     & 0.037 & 0     & 0.016 \\
        SST-2       & 0.061 & 0.016 & 0.035  \\
        AG-News     & 0.009 & 0.013 & 0.013 \\
        \bottomrule
        \textbf{Accuracy} & & \\
        SST-2       & 0.035 & 0.022  & 0.033 \\
        AG-News     & 0.004 & 0.004  & 0.004 \\
        \bottomrule
    \end{tabular}
    \caption{Standard deviations of PLMs' performance under different types of noise.}
    \label{tab:std_by_noise}
\end{table}

\end{document}


\onecolumn
\appendix
\section{Noisy Sentences} \label{app:noisy_sentences}

\begin{table*}[h!]
    \centering
    \begin{tabular}{llp{12cm}}
        Dataset & Type & Examples  \\
         \hline
         \hline
         \multirow{2}{*}{SST-2} 
         & Clean & The story may not be new, but Australian director John Polson, making his American feature debut, jazzes it up adroitly. 
        \\
        \cmidrule{2-3}
        & Noisy & The story \textcolor{red}{\textit{\textbf{mwy}}} not be new, \textcolor{red}{\textit{\textbf{bt}}} Australian \textcolor{red}{\textit{\textbf{directlr}}} John Polson, making his American feature \textcolor{red}{\textit{\textbf{deubt}}}, \textcolor{red}{\textit{\textbf{jazezs}}} it up \textcolor{red}{\textit{\textbf{adrooitly}}}.
        \\
        \midrule
        \multirow{2}{*}{Yelp} 
         & Clean & the salsa is great, the fish top notch, the drinks just excellent.
        \\
        \cmidrule{2-3}
        & Noisy & the \textcolor{red}{\textit{\textbf{slsa}}} is great, the \textcolor{red}{\textit{\textbf{fsh tpp nootch}}}, the drinks \textcolor{red}{\textit{\textbf{juxt excellegt}}}. 
        \\
        \midrule
        \multirow{2}{*}{AG-News} 
         & Clean & Fears for T N pension after talks Unions representing workers at Turner Newall say they are' disappointed' after talks with stricken parent firm Federal Mogul.
        \\
        \cmidrule{2-3}
        & Noisy & Fears for T N pension after talks Unions representing workers at \textcolor{red}{\textit{\textbf{Tunrer}}} Newall \textcolor{red}{\textit{\textbf{saay tehy ate' dksappointed' aftr}}} talks with stricken parent firm \textcolor{red}{\textit{\textbf{Fexeral}}} Mogul.
        \\
        \bottomrule

    \end{tabular}
    \caption{Examples of Noisy Sentences}
     \label{tab:noisy_sentences}
\end{table*}



\section{Model Performance of Word Corruption} \label{app:intact_wc}

\begin{table*}[h!]
    \centering
    \begin{tabular}{l|r|r|r}
        \toprule
        Models    &  Intact   & Partial & Failed  \\
        \midrule
        BERT      &   \textbf{0.56} & 0.64 &  0.78\\
        RoBERTa   &   \textbf{0.94} & 0.96 & 0.98\\
        ALBERT    &   \textbf{0.54} & 0.79 & 0.89\\
        \bottomrule
    \end{tabular}
    \caption{Performance of PLMs under different types of word corruption. 
    }
    \label{tab:types_corrupt}



\end{table*}
The table evaluates the performance of PLMs under different types of word corruption. The less the value, the worse the performance.
The \emph{intact word corruption}, as we expect, is most likely to cause the mis-understanding of PLMs. Table \ref{tab:types_corrupt} illustrates the result.

        
    
\section{Performance of PLMs on Samples with Different M-scores} \label{app:plm-misc-sim}
Since the averaged M-scores for noisy sentences would be a continuous value, we group noisy samples according to rounded M-scores and then calcuate the metrics on each group. It shows that, for noisy words, the higher M-scores the groups have, the more poorly the models would perform. However, after a certain threshold (e.g., $\text{M-scores} > 4$ for SST-2), the performance of noisy sentences becomes better when their M-scores become larger.




\begin{table*}[h!]
    \centering
    \begin{tabular}{llrrrrrr}
        \toprule
        \multirow{2}{*}{Data} & \multirow{2}{*}{WCR-1} & \multicolumn{2}{c}{BERT} &
        \multicolumn{2}{c}{RoBERTa}&
        \multicolumn{2}{c}{ALBERT}\\
        \cmidrule{3-8}
         & & Misc. & Sent Sim
           & Misc. & Sent Sim 
           & Misc. & Sent Sim  \\
        \midrule
        
        \multirow{3}{*}{SST-2}
        & 1  & 0.06  & 0.94
             & 0.10  & 0.86
             & 0.05  & 0.91 
             \\
        & 2  & 0.12 & 0.87
             & 0.14 & 0.82
             & 0.11 & 0.85 
             \\
        & 3  & 0.16 & 0.83
             & 0.10 & 0.86
             & 0.16 & 0.81
             \\
        & 4  & 0.12 & 0.89
             & 0    & 0.9
             & 0.09 & 0.88
             \\
        & 5  & / & /
             & / & /
             & 0 & 1
             \\
        \midrule
        \multirow{3}{*}{Yelp}
        & 1  & 0.17 & 0.83
             & 0.13 & 0.85
             & 0.10 & 0.89
             \\
        & 2  & 0.30 & 0.70
             & 0.26 & 0.74
             & 0.27 & 0.71
             \\
        & 3  & 0.26 & 0.74
             & 0.21 & 0.79
             & 0.28 & 0.71
             \\
        & 4  & 0.17 & 0.83
             & 0.05 & 0.93
             & 0.24 & 0.75
             \\
        \midrule
        \multirow{3}{*}{AG-News}
        & 1  & 0.03 & 0.96
             & 0    & 1
             & 0    & 0.98
             \\
        & 2  & 0.05 & 0.95
             & 0.10 & 0.89
             & 0.06 & 0.93 
             \\
        & 3  & 0.05 & 0.94
             & 0.04 & 0.86
             & 0.06 & 0.93 
             \\
        & 4  & 0.11 & 0.9
             & 0    & 1
             & 0.02 & 0.95
             \\
        
        \midrule
        \multirow{3}{*}{Setiment Lexicon}
        & 1  & / & 0.78
             & / & 0.97
             & / & 0.75 \\
        & 2  & / & 0.73
             & / & 0.96
             & / & 0.55\\
        & 3  & / & 0.64
             & / & 0.95
             & / & 0.30 \\
        & 4  & / & 0.56
             & / & 0.94
             & / & 0.17 \\
        & 5  & / & 0.45
             & / & /
             & / & 0.02 \\
        \bottomrule
    \end{tabular}
    \caption{Model performance on noisy samples with different M-scores.\textit{Sent Sim} represents the average similarity between the noisy samples and their corresponding clean texts, while \textit{Misc.} indicates the misclassification rates. }
    \label{tab:main}
\end{table*}